\documentclass{article}

\usepackage{microtype}
\usepackage{graphicx}
\usepackage{subcaption}
\usepackage{booktabs} 

\usepackage{hyperref}

\usepackage[accepted]{icml2026}

\usepackage{amsmath}
\usepackage{amssymb}
\usepackage{mathtools}
\usepackage{amsthm}
\usepackage{bm}
\usepackage{multirow}
\usepackage{colortbl}

\usepackage[capitalize,noabbrev]{cleveref}

\theoremstyle{plain}

\theoremstyle{definition}

\theoremstyle{remark}

\newcommand{\ie}{\textit{i}.\textit{e}., }
\newcommand{\eg}{\textit{e}.\textit{g}. }

\usepackage[textsize=tiny]{todonotes}
\usepackage{etoolbox}

\hypersetup{breaklinks=false}
\icmltitlerunning{VersatileFFN}

\begin{document}
	
	\twocolumn[
	\icmltitle{VersatileFFN: Achieving Parameter Efficiency in LLMs via Adaptive Wide-and-Deep Reuse}
	
	
	
	\icmlsetsymbol{equal}{*}
	
	\begin{icmlauthorlist}
		\icmlauthor{Ying Nie}{comp}
		\icmlauthor{Kai Han}{comp}
		\icmlauthor{Hongguang Li}{comp}
		\icmlauthor{Hang Zhou}{comp}
		\icmlauthor{Tianyu Guo}{comp}
		\icmlauthor{Enhua Wu}{sch1,sch2}
		\icmlauthor{Xinghao Chen}{comp}
		\icmlauthor{Yunhe Wang}{comp} \\
	\end{icmlauthorlist}
	\icmlaffiliation{comp}{Huawei Noah’s Ark Lab}
	\icmlaffiliation{sch1}{ISCAS}
	\icmlaffiliation{sch2}{University of Macau}
	
	\icmlcorrespondingauthor{Kai Han}{kai.han@huawei.com}
	\icmlcorrespondingauthor{Yunhe Wang}{yunhe.wang@huawei.com}
	
	\icmlkeywords{Machine Learning, ICML}
	
	\vskip 0.3in
	]
	
	
	
	\printAffiliationsAndNotice{}  
	
	\begin{abstract}
		The rapid scaling of Large Language Models (LLMs) has achieved remarkable performance, but it also leads to prohibitive memory costs. Existing parameter‑efficient approaches such as pruning and quantization mainly compress pretrained models without enhancing architectural capacity, thereby hitting the representational ceiling of the base model. In this work, we propose VersatileFFN, a novel feed‑forward network (FFN) that enables flexible reuse of parameters in both width and depth dimensions within a fixed parameter budget. Inspired by the dual‑process theory of cognition, VersatileFFN comprises two adaptive pathways: a width‑versatile path that generates a mixture of sub‑experts from a single shared FFN, mimicking sparse expert routing without increasing parameters, and a depth‑versatile path that recursively applies the same FFN to emulate deeper processing for complex tokens. A difficulty‑aware gating dynamically balances the two pathways, steering “easy” tokens through the efficient width‑wise route and allocating deeper iterative refinement to “hard” tokens. Crucially, both pathways reuse the same parameters, so all additional capacity comes from computation rather than memory. Experiments across diverse benchmarks and model scales demonstrate the 
		effectiveness of the method. The code is available at \url{https://github.com/huawei-noah/noah-research/tree/master/VersatileFFN}.
	\end{abstract}

	\section{Introduction}
	\begin{figure*}[t]
		\begin{center}
			\centering
			\includegraphics[width=0.95\linewidth]{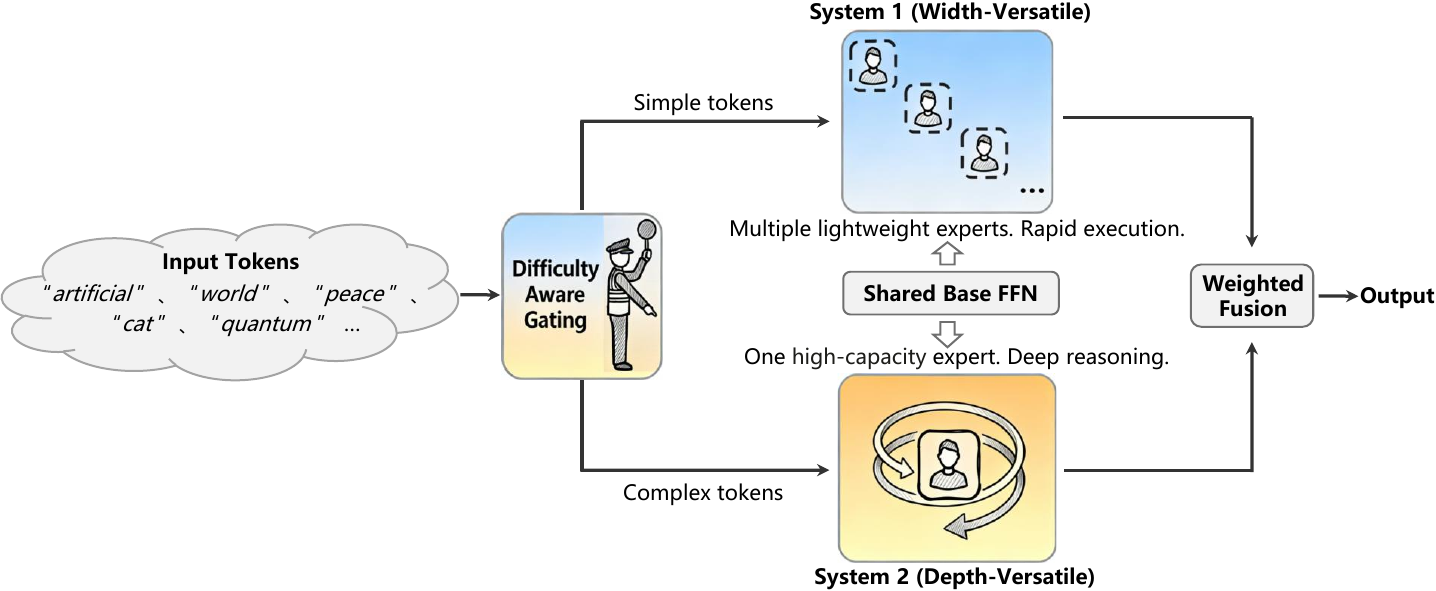}
			\caption{VersatileFFN integrates a width-versatile pathway for rapid execution and a depth-versatile pathway for deep reasoning, both derived from a shared base FFN. This design enables a flexible trade-off between the two computational dimensions.}
			\label{fig:intro}
		\end{center}
		\vspace{-1.0em}
	\end{figure*}
	
	The remarkable success of Large Language Models (LLMs) is largely governed by scaling laws, which posit a power-law relationship between model performance and parameter count~\cite{kaplan2020scaling, hoffmann2022training}. Following this trajectory, the pursuit of state-of-the-art accuracy has led to the proliferation of models with hundreds of billions of parameters. This trend encompasses both dense architectures~\cite{touvron2023llama, chen2025pangu, zuo2025man} and sparse Mixture-of-Experts (MoE) variants~\cite{jiang2024mixtral, dai2024deepseekmoe, comanici2025gemini, yang2025qwen3, team2025kimi}, reflecting a consistent drive toward increasing model scale.
	
	However, the massive parameter counts of these models incur a prohibitive memory footprint that severely limits their practical deployment. High-end accelerators with large memory capacity are expensive and scarce, and distributing model weights across multiple devices introduces significant communication overhead and engineering complexity. As a result, scaling models purely by increasing parameter counts faces growing infrastructural and economic constraints.
	
	In recent years, a line of research has explored parameter-efficient LLM architectures that maintain high performance under limited memory. Such methods include weight pruning~\cite{frantar2023sparsegpt,chen2025pangu}, quantization~\cite{xiao2023smoothquant,fu2025eaquant}, low-rank adaptations~\cite{lu2025flrc}, and other compression techniques that reduce the storage footprint of pretrained models. However, these approaches primarily aim to approximate the capabilities of the original large model, and thus remain fundamentally constrained by its architectural capacity. They do not augment the model’s representational power, nevertheless, they trade off precision or connectivity for deployability, often hitting an upper bound determined by the base model’s design. There remains a clear need for architectures that are parameter-efficient by design, capable of attaining stronger performance under the same parameter budget, rather than merely compressing existing large models.
	
	In this work, we introduce VersatileFFN, a parameter-efficient architecture that versatilely reuse parameters and computation in feed-forward networks (FFNs), serving as an efficient replacement for the standard FFN in Transformer blocks while leaving the self-attention component unchanged. As illustrated in Figure~\ref{fig:intro}, VersatileFFN integrates two complementary mechanisms: a width-versatile pathway and a depth-versatile pathway, which together enable flexible trade-offs between computational width and depth using a tightly shared parameter set.
	\begin{itemize}
		\item In the width-versatile FFN pathway, we introduce the concept of a versatile base expert, a single and multi-capable FFN, from which we derive a mixture of sub-experts through structured parameter reuse. Rather than instantiating separate experts, each sub-expert is formed by selectively activating and composing different, non-overlapping hidden subspaces of the same underlying versatile FFN. This approach enables the emulation of diverse expert behaviors with minimal memory overhead, preserving the adaptive token routing benefits of classical MoE architectures while avoiding the parameter proliferation typical of such designs.
		\item The depth-versatile FFN pathway achieves depth-wise versatility by recursively reusing the same base FFN multiple times, effectively emulating the function of multiple layers. Rather than instantiating separate parameters for each processing step, this design enables tokens to undergo iterative refinement through cyclical application of the shared MLP. A Gumbel-Softmax-based controller predicts a token-specific iteration count, allowing the model to dynamically allocate more processing steps to harder tokens. This approach maintains full differentiability during training while preserving parameter efficiency, as the same weights are repurposed depth-wise to create a versatile, adaptive computation structure.
	\end{itemize}
	Inspired by the dual-process theory of human cognition~\cite{kahneman2011thinking}, we design an architecture that dynamically allocates computation: “easy” tokens are processed rapidly via a lightweight path, while “hard” tokens receive deeper, iterative reasoning, all under a fixed parameter budget. A difficulty-aware gating mechanism unifies the two pathways, using the expected iteration count from the depth-versatile controller as a proxy for token difficulty. This signal modulates the fusion weights between the outputs of width-versatile FFN and depth-versatile FFN, favoring efficient processing on width-versatile for easy tokens and shifting toward depth-versatile for harder ones. Crucially, both pathways share the same MLP parameters, ensuring added capacity comes from computation, not memory.
	We conduct extensive experiments across multiple  benchmarks and model scales. Models equipped with VersatileFFN consistently outperform other parameters-matched or FLOPs-matched methods, demonstrating the effectiveness of our dual-process design.

	\section{Related Work}
	\subsection{Mixture-of-Experts}
	The Mixture-of-Experts (MoE) architecture serves as a foundational framework for scaling model capacity without a proportional increase in computational cost. Originally introduced by~\cite{jacobs1991adaptive, jordan1994hierarchical}, the concept was later popularized in the large-scale networks by~\cite{shazeer2017outrageously}. Subsequent advancements, such as GShard~\cite{lepikhin2020gshard} and Switch Transformer~\cite{fedus2022switch}, have demonstrated that replacing standard feed-forward layers in Transformers with MoE layers facilitates efficient pre-training at the trillion-parameter scale, yielding substantial performance improvements.
	Recent advancements have focused on refining expert granularity and routing strategies~\cite{jiang2024mixtral, dai2024deepseekmoe, jin2024moe++, huang2024harder, wang2024remoe}, enabling the successful deployment of large-scale MoE models in industrial applications~\cite{yang2025qwen3, comanici2025gemini, team2025kimi, zhao2025towards}.
	
	Despite their inference efficiency, the massive storage requirements of standard MoE models present significant deployment hurdles. This has catalyzed research into parameter-efficient MoE, utilizing techniques such as expert merging~\cite{li2022branch, zhao2024hypermoe}, pruning~\cite{sarkar2024revisiting, lu2024not}, and quantization~\cite{nie2022redistribution, dong2024stbllm, huang2024mixture, zhou2025floe}. Another emerging direction involves composing lightweight, task-specific modules (\eg LoRA adapters) into mixture-based systems, as seen in LoRAHub~\cite{huang2023lorahub}, MoA~\cite{feng2024mixture},  MoLE~\cite{wu2024mixture}, and MoRAgent~\cite{han2025moragent}. While these approaches mitigate some memory constraints, they typically rely on instantiating distinct, physically separate expert parameters, which limits the extent of parameter efficiency compared to architectures designed for intrinsic reuse.
	
	\subsection{Recursive Computation and Adaptive Inference}
	A promising avenue for maximizing parameter efficiency is the decoupling of model depth from parameter count through recursive computation. Early works such as Universal Transformers~\cite{dehghani2018universal} and ALBERT~\cite{lan2019albert} demonstrate that cross-layer parameter sharing can induce beneficial inductive biases and improve parameter efficiency for
	language modeling. Recent theoretical analysis further establishes that such recursive architectures can emulate complex algorithms, acting as universal computers~\cite{giannou2023looped, gao2024expressive}, and generalize to sequence lengths far beyond those encountered during training~\cite{fan2024looped, gong2025makes}.
	
	Beyond the paradigm of static parameter sharing, recursive structures facilitate dynamic computation, where the computational budget is adapted to the complexity of the input. Recent advancements such as Mixture-of-Depths~\cite{raposo2024mixture}, Mixture-of-Recurrence~\cite{bae2025mixture} and Dynamic resolution network~\cite{zhu2021dynamic} dynamically allocate inference FLOPs, allowing models to expend more "thinking time" on harder tokens or images. This iterative refinement is particularly potent for reasoning tasks, offering significant advantages over static counterparts. Several studies~\cite{gatmiry2024can, saunshi2025reasoning, merrill2025little, zhu2025scaling} indicate that increasing computational depth often yields greater performance gains than merely increasing width. In addition, HRM~\cite{wang2025hierarchical} and TRM~\cite{jolicoeur2025less} show that iteratively reasoning models can outperform larger static counterparts. 
	While these approaches successfully exploit the depth dimension for efficiency, they typically treat the recurrent layer as a monolithic block, leaving the potential for fine-grained, width-wise parameter reuse largely unexplored. Ours aims to bridge this gap by introducing a multi-dimensional sharing mechanism that exploits redundancy across both depth and width.

	\section{Method}
	
	\subsection{Architectural Overview}
	Let $\mathbf{X} \in \mathbb{R}^{B \times T \times d}$ denote the input tensor to the Transformer block, where $B$ represents the batch size, $T$ the sequence length and $d$ the feature dimension.  In a standard Transformer layer, the input first undergoes a Self-Attention mechanism followed by residual connection and normalization:
	\begin{equation}
	\mathbf{H} = \mathbf{X} + \text{Attention}(\text{LayerNorm}(\mathbf{X})),
	\label{eq:att}
	\end{equation}
	Subsequently, the hidden states are processed by a Feed-Forward Network (FFN). We define the FFN transformation function $\mathcal{F}(\cdot)$ as:
	\begin{equation}
	\begin{aligned}
	\mathbf{Y} &= \mathcal{F}(\mathbf{H}) \\
	&=\mathbf{H} + \mathbf{W}_{out} \, \phi(\mathbf{W}_{proj} \,\text{LayerNorm}(\mathbf{H})).
	\label{eq:ffn}
	\end{aligned}
	\end{equation}
	where $\mathbf{W}_{proj} \in \mathbb{R}^{d \times d_{hidden}}$, $\mathbf{W}_{out} \in \mathbb{R}^{d_{hidden} \times d}$ denote the projection and output weights, respectively, and $\phi$ is a distinct non-linear activation function.
	
	We introduce VersatileFFN, a parameter-efficient alternative to the standard FFN. While retaining the canonical self-attention architecture, VersatileFFN reconfigures the feed-forward computation into two complementary pathways that share the underlying FFN weights (\ie $\mathbf{W}_{proj}$ and $\mathbf{W}_{out}$):
	\begin{itemize}
		\item Width-Versatile ($\mathbf{Y}_{width}$): This pathway functions as a virtual MoE module. It routes tokens to specialized sub-experts instantiated via structured subsets of the shared weights, facilitating rapid, domain-specialized response without increasing parameter count.
		
		\item Depth-Versatile ($\mathbf{Y}_{depth}$): This pathway implements a recursive computation mechanism. It iteratively refines token representations by reusing the full capacity of $\mathbf{W}_{proj}$ and $\mathbf{W}_{out}$, thereby allocating increased computational depth to semantically complex tokens.
	\end{itemize}
	
	The final output $\mathbf{Y}$ is synthesized through a dynamic, difficulty-aware fusion of these two pathways:
	\begin{equation}
	\mathbf{Y} = \lambda \cdot \mathbf{Y}_{width} + (1 - \lambda) \cdot \mathbf{Y}_{depth},
	\label{eq:fusion}
	\end{equation}
	where $\lambda \in [0,1)$ acts as a gating coefficient modulated by the token difficulty. Specifically, $\lambda$ is derived from the predicted iteration count of the depth-versatile controller. This mechanism ensures a flexible computational trade-off: ensuring broad semantic coverage for simple tokens via the width pathway, while reserving deep, iterative reasoning for harder tokens via the depth pathway, all within a fixed parameter budget.

	\subsection{Width-Versatile Mechanism}
	The Width-Versatile pathway, implemented as a virtual MoE, augments the model's representational capacity while circumventing the prohibitive memory overhead typically associated with physical expert instantiation.
	
\begin{figure}[t]
		\begin{center}
			\centering
			\includegraphics[width=0.95\linewidth]{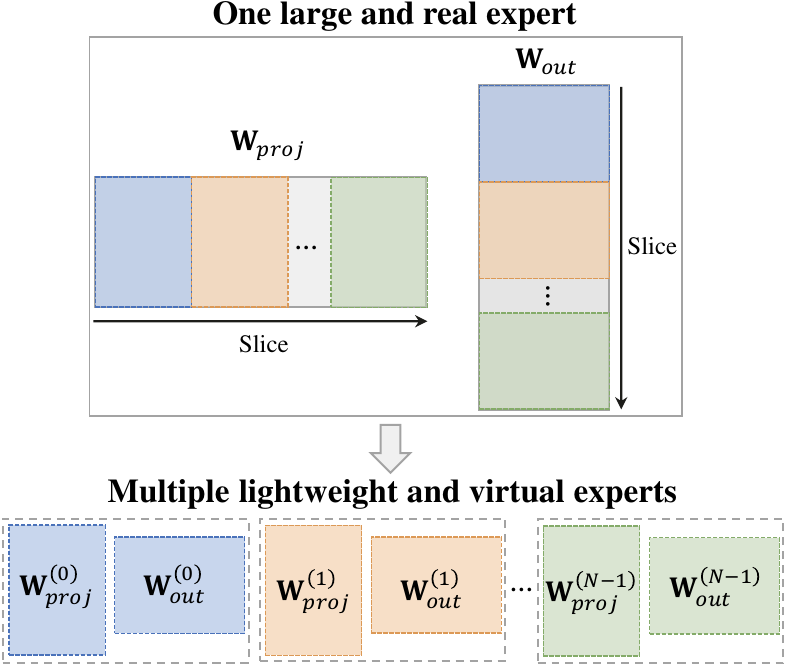}
			\caption{Illustration of decoupling one large and real expert into multiple lightweight and virtual experts.}
			\label{fig:method}
		\end{center}
		\vspace{-1.0em}
	\end{figure}

	\noindent\textbf{Construction of Virtual Experts. }
	Rather than allocating discrete weight matrices for each expert, we leverage the shared weights $\mathbf{W}_{proj}$ and $\mathbf{W}_{out}$ as a contiguous parameter substrate. We define a set of $N$ virtual experts by extracting structured, non-overlapping subspaces from the hidden dimension $d_{hidden}$ as illustrated in Figure~\ref{fig:method}. Specifically, let $d_{expert}$ denote the hidden dimension of a single virtual expert. To strictly satisfy the non-overlapping requirement, we impose the constraint $N \times d_{expert} \le d_{hidden}$, ensuring that the total capacity of the virtual experts does not exceed the shared parameter space. We employ a strided slicing strategy to map each expert $k \in \{0, \dots, N-1\}$ to a specific view of the shared parameters. This design adheres to two principles: (1) \textit{Parameter Efficiency}, achieved through the dense reuse of the backbone weights. (2) \textit{Functional Orthogonality}, ensured by the non-overlapping allocation, which prevents interference between experts. 
	
	The stride $S$ is calculated to uniformly distribute expert views across the hidden dimension:
	\begin{equation}
	S = \left\lfloor \frac{d_{hidden} - d_{expert}}{N - 1} \right\rfloor.
	\label{eq:step}
	\end{equation}
	The indices $\mathcal{I}_{proj}^{(k)}$ corresponding to the $k$-th expert is:
	\begin{equation}
	\mathcal{I}_{proj}^{(k)} = \{j \in \mathbb{Z} \mid \lfloor k \cdot S \rfloor \le j < \lfloor k \cdot S \rfloor + d_{expert} \}.
	\label{eq:index}
	\end{equation}
	Leveraging the pre-trained weights of the dense model, we assign $\mathcal{I}_{out}^{(k)} = \mathcal{I}_{proj}^{(k)}$ to maintain the structural alignment between the projection and output. The effective weight matrices for the virtual expert $k$ are thus derived as:
	\begin{equation}
	\mathbf{W}_{proj}^{(k)} = \mathbf{W}_{proj}[:, \mathcal{I}_{proj}^{(k)}], \quad \mathbf{W}_{out}^{(k)} = \mathbf{W}_{out}[\mathcal{I}_{out}^{(k)}, :].
	\label{eq:build_moe}
	\end{equation}
	
	\noindent\textbf{Sparse Token Routing. }
	We employ a learnable gating to orchestrate token assignment. Given the post‑attention token representation $\mathbf{H} \in \mathbb{R}^{B \times T \times d}$, a router $\mathbf{W}g \in \mathbb{R}^{d \times N}$ computes the gating logits $G(\mathbf{H}) = \mathbf{W}_g \mathbf{H}$. We adopt a standard Top-$K$ strategy, activating only the experts with the highest routing scores. The output $\mathbf{Y}_{width}$ is computed as the probability-weighted sum of the selected experts:
	\begin{equation}
	\mathbf{Y}_{width} = \sum_{k \in \text{Top-}K(G(\mathbf{H}))} g_k \cdot \mathbf{Y}_{k},
	\label{eq:y_moe}
	\end{equation}
	where $g_k$ represents the Softmax-normalized gating probability, and the expert-specific transformation $\mathbf{Y}_{k}(\mathbf{H})$ is:
	\begin{equation}
	\mathbf{Y}_{k} = \mathbf{H} + {\mathbf{W}}_{out}^{(k)} \, \phi({\mathbf{W}}_{proj}^{(k)} \,\text{LayerNorm}(\mathbf{H})).
	\label{eq:y_moe_single}
	\end{equation}
	This sparse activation mechanism decouples computational cost from model capacity. Following previous experience~\cite{fedus2022switch}, we incorporate an auxiliary load-balancing loss during training to prevent expert collapse.
	
	\subsection{Depth-Versatile Mechanism}
	
	The Depth-Versatile pathway introduces token-wise adaptive computation by applying the shared MLP block recursively, enabling dynamic depth allocation.
	
	\noindent\textbf{Recursive Weight Application. }
	In contrast to the Width-Versatile pathway, The Depth-Versatile pathway utilizes the entire shared backbone weights ($\mathbf{W}_{proj}, \mathbf{W}_{out}$) without slicing. This ensures that the recursive refinement process benefits from the full expressivity of the model parameters. Let $\mathcal{F}(\cdot)$ denote the standard FFN transformation defined in Eq.~\ref{eq:ffn}. We generate a sequence of intermediate representations by recursively applying $\mathcal{F}$:
	\begin{equation}
	\mathbf{H}^{(\ell)} = \mathcal{F}\big(\mathbf{H}^{(\ell-1)}\big),\quad \ell = 1,\dots,L_{max},
	\label{eq:loop}
	\end{equation}
	where $\mathbf{H}^{(0)} = \mathbf{H}$, and \(L_{max}\) denotes the maximum allowable iterations. This iterative process allows the model to progressively refine token representations, capturing complex dependencies through repeated non-linear transformations without increasing the parameter budget.
	
	\noindent\textbf{Differentiable Loop Prediction. }
	The optimal number of iterations per token is inherently discrete and non-differentiable. To facilitate end-to-end training, we introduce a prediction head $\mathbf{W}_{loop} \in \mathbb{R}^{d \times L_{max}}$ to estimate the required computational depth. The logits for the loop count are computed as $P_{loop}(\mathbf{H}) = \mathbf{W}_{loop} \mathbf{H}$. We then employ the Gumbel-Softmax relaxation~\cite{jang2016categorical} to sample a differentiable probability vector $\mathbf{p} \in \Delta^{L_{max}-1}$:
	\begin{equation}
	\mathbf{p} = \text{Softmax}\left( \frac{P_{loop}(\mathbf{H}) + \mathbf{g}}{\tau} \right),
	\label{eq:gumbel}
	\end{equation}
	where $\mathbf{g} \sim \text{Gumbel}(0, 1)$ and $\tau$ is the temperature. 
	
	We utilize the Straight-Through Estimator (STE) during training: the forward pass executes a discrete selection (via \textit{argmax}) to simulate the inference-time decision, while gradients are propagated through the continuous relaxation $\mathbf{p}$ during the backward pass. To stabilize convergence, $\tau$ is annealed exponentially from an initial value (\eg 5.0) to a lower bound (\eg 0.1). The output of the Depth-Versatile pathway is computed as the soft-weighted combination of intermediate states: 
	\begin{equation}
	\mathbf{Y}_{depth} = \sum_{\ell=1}^{L_{max}} \mathbf{p}_{\ell} \cdot \mathbf{H}^{(\ell)}.
	\label{eq:y_loop}
	\end{equation}
	During inference, the probabilistic sampling is replaced by a deterministic decision $\hat{\ell} = \arg\max(\mathbf{p})$. The loop is executed exactly $\hat{\ell}$ times, and the final state is taken as $\mathbf{Y}_{depth} = \mathbf{H}^{(\hat{\ell})}$.
	
	\subsection{Difficulty-Aware Fusion}
	
	The Width-Versatile and Depth-Versatile pathways represent distinct yet complementary computational paradigms: the former offers broad, parallelizable semantic capacity via virtual experts, while the latter facilitates intensive, sequential reasoning through iterative refinement. VersatileFFN synergizes these mechanisms via a difficulty-aware fusion scheme, which leverages the predicted computational depth as a proxy for token complexity.
	
	\noindent\textbf{Expected Loop Count as Difficulty Proxy. }
	We postulate that the requisite depth of recursive processing serves as an intrinsic measure of semantic difficulty. Linguistically trivial tokens (\eg stopwords, high-frequency bigrams) typically necessitate minimal transformation (loop count $\to 1$), whereas semantically ambiguous or logically complex tokens benefit from deep, recurrent refinement (loop count $\to L_{max}$).
	
	We quantify this difficulty by computing the expected loop count, $\mathbb{E}[L]$, derived from the soft probability distribution $\mathbf{p}$ output by the loop predictor:
	\begin{equation}
	\mathbb{E}[L] = \sum_{\ell=1}^{L_{max}} \ell \cdot \mathbf{p}_{\ell}.
	\label{eq:expect_loop}
	\end{equation}
	During training, the differentiability of $\mathbf{p}$ allows $\mathbb{E}[L] \in [1, L_{max}]$ to serve as a continuous signal for gradient propagation. During inference, although the loop execution becomes discrete, $\mathbb{E}[L]$ remains a robust indicator of the model's uncertainty and the token's processing demand.
	
	\begin{algorithm}[t]
		\caption{Training Computation of VersatileFFN}
		\label{alg:versatile_train}
		\begin{algorithmic}[1]
			\REQUIRE Input tensor $\mathbf{X}$, Shared Weights $\mathbf{W}_{proj}, \mathbf{W}_{out}$, Max loops $L_{max}$
			\ENSURE Output tensor $\mathbf{Y}$
			
			\STATE Compute post-attention representation $\mathbf{H}$ via Eq.~\ref{eq:att}
			
			\STATE \textbf{\textit{Width-Versatile Pathway:}}
			\STATE \hspace{1em} Construct virtual experts by slicing shared weights via Eq.~\ref{eq:index} and Eq.~\ref{eq:build_moe}
			\STATE \hspace{1em} Compute output $\mathbf{Y}_{width}$ with Top-$k$ routing via Eq.~\ref{eq:y_moe}
			
			\STATE \textbf{\textit{Depth-Versatile Pathway:}}
			\STATE \hspace{1em} Predict loop count probabilities $\mathbf{p}$ via Eq.~\ref{eq:gumbel}
			\STATE \hspace{1em} Initialize $\mathbf{H}^{(0)} \leftarrow \mathbf{H}$
			\STATE \hspace{1em} \textbf{for} $\ell = 1$ to $L_{max}$ \textbf{do}
			\STATE \hspace{2em} Update recursive hidden state $\mathbf{H}^{(\ell)}$ using full shared weights via Eq.~\ref{eq:loop}
			\STATE \hspace{1em} \textbf{end for}
			\STATE \hspace{1em} Aggregate states to obtain $\mathbf{Y}_{depth}$ via Eq.~\ref{eq:y_loop}
			
			\STATE \textbf{\textit{Difficulty-Aware Fusion:}}
			\STATE \hspace{1em} Calculate expected loop count $\mathbb{E}[L]$ via Eq.~\ref{eq:expect_loop}
			\STATE \hspace{1em} Compute difficulty-aware fusion scalar $\lambda$ via Eq.~\ref{eq:fusion_weights}
			\STATE \hspace{1em} Compute output $\mathbf{Y}$ by fusing pathways via Eq.~\ref{eq:fusion}
			
			\STATE \textbf{Return} $\mathbf{Y}$
		\end{algorithmic}
	\end{algorithm}
	
	\noindent\textbf{Dynamic Fusion Modulation. }
	To unify the two pathways, we define a dynamic gating scalar, $\lambda$, which modulates the fusion balance based on the estimated difficulty. We formulate $\lambda$ to be inversely proportional to the expected computational cost:
	\begin{equation}
	\lambda = \frac{L_{max} - \mathbb{E}[L]}{L_{max}}.
	\label{eq:fusion_weights}
	\end{equation}
	
	By construction, $\lambda \in [0,1)$. For ``easy'' tokens with low expected loops, $\lambda \to 1$, biasing the output toward the efficient width. Conversely, for ``hard'' tokens, $\lambda \to 0$ , shifting the focus toward the depth. The final output $\mathbf{Y}$  is synthesized according to Eq.~\ref{eq:fusion}. This mechanism automatically allocates computational resources: simple patterns are resolved rapidly via the lightweight virtual MoE, while complex reasoning tasks command the full depth of the recursive loop. 
	
	\noindent\textbf{Inference Optimization. }
	The complete training procedure is summarized in Algorithm~\ref{alg:versatile_train}. To minimize latency during inference, we introduce two inference-time optimizations:
	\begin{itemize}
		\item Discrete Early-Exit: The Depth-Versatile pathway transitions from soft aggregation to a hard cutoff. Recursion terminates immediately at the predicted step  $\hat{\ell} = \arg\max(\mathbf{p})$, avoiding unnecessary computation beyond the required depth.
		\item Conditional Parallelism: We employ a threshold-based execution strategy. If the contribution of the Width-Versatile pathway is negligible (\ie $\lambda = 0$) the branch is pruned entirely. Otherwise, both pathways are executed in parallel to maximize hardware utilization and computational throughput.
	\end{itemize}
	The inference procedure is summarized in Algorithm~\ref{alg:versatile_infer}.
	
	\section{Experiments}
	\subsection{Implementation Details}
	\noindent\textbf{Experiments Setup. } 
	To validate our approach, we conducted pre-training experiments on the FineWeb-Edu dataset~\cite{penedo2024fineweb}, our implementation is based on OLMo2~\cite{olmo20242}. Specifically, we train a 354M parameter model on 40B tokens, a 720M parameter model on 70B tokens and a 1.21B parameter model on 100B tokens. The 40B and 70B token sets are subsets derived from the 100B token corpus~\cite{huggingfacefw_2024}. Following the architectural settings of the open-source OLMo2 models, we set the hidden dimensions to 1,024 for 354M model, 1,536 for the 720M model and 2,048 for 1.21B model. All models consist of 15 layers and utilize SwiGLU for activation and RMSNorm~\cite{zhang2019root} for normalization. Besides, The OLMo2 tokenizer with a vocabulary size of 50,280 is employed in our models. Further training setting and evaluation benchmarks are provided in Appendix~\ref{appendix_0}.
	
	\begin{figure}[t] 
		\centering
		\begin{subfigure}{0.48\textwidth}
			\includegraphics[width=\linewidth]{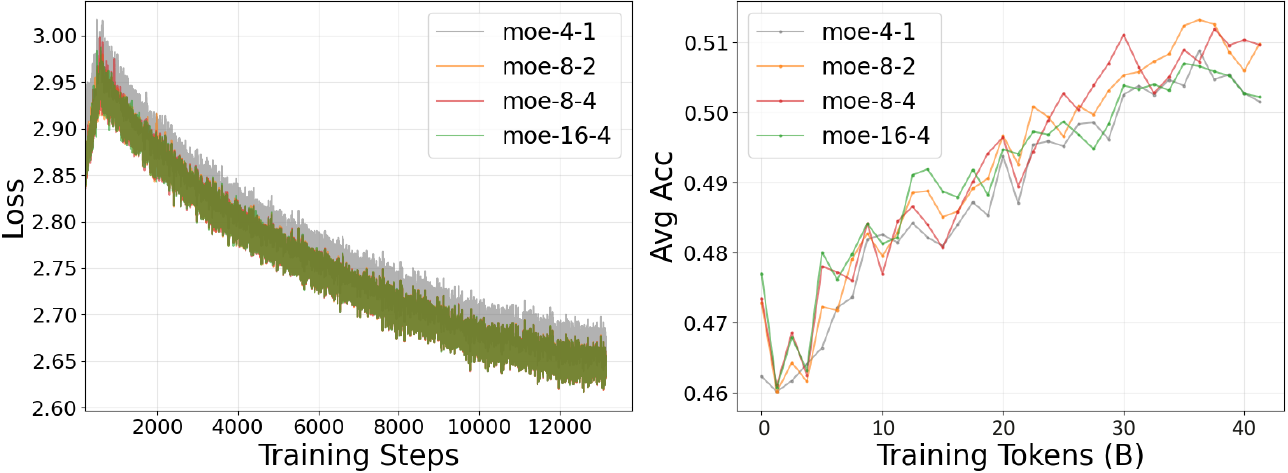}
			\caption{Various total experts and active experts.}
			\label{fig:top}
		\end{subfigure}
		\begin{subfigure}{0.48\textwidth}
			\includegraphics[width=\linewidth]{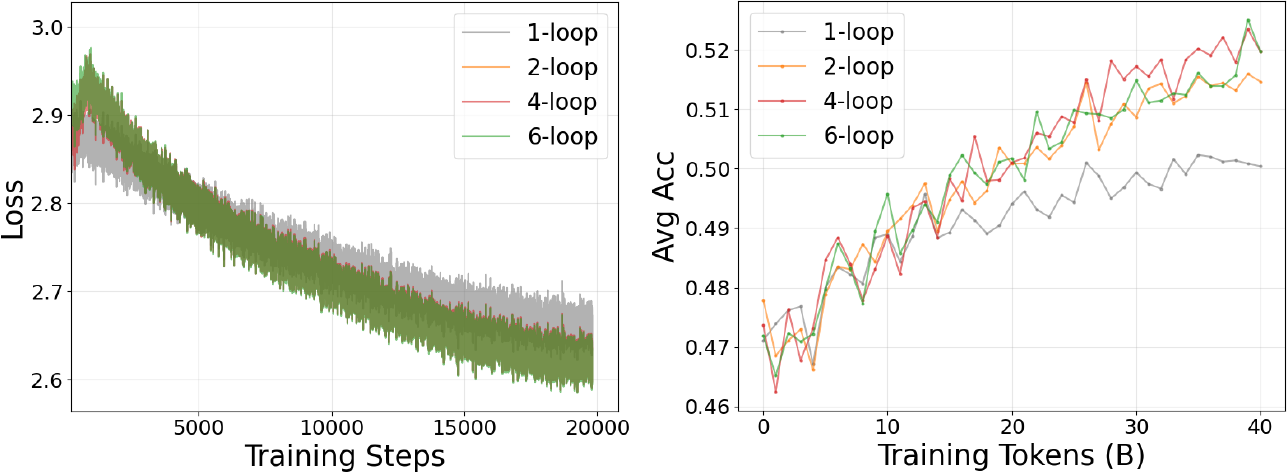}
			\caption{Various maximum number of loops.}
			\label{fig:bottom}
		\end{subfigure}
		\caption{The curves of training loss and average accuracy for various configurations of width-versatile and depth-versatile mechanisms.}
		\label{fig:exp_settting}
		\vspace{-1.0em}
	\end{figure}

	\begin{table*}[t]
		\caption{Comparison of zero-shot performance on standard NLP benchmarks. Results are reported for two model scales. "Avg." indicates the average accuracy across all eight tasks.}
		\label{table:main_result}
		\vspace{-1em}
		\begin{center}
			\begin{small}
				\begin{sc}
					\begin{tabular}{c|c|ccccccccccc}
						\toprule
						Method  & Loss & PIQA & HellaSwag &OBQA & SciQ & ARC-e & ARC-c & COMM & WINO &Avg. \\
						\midrule
						\multicolumn{11}{c}{\cellcolor[HTML]{EFEFEF}\textit{Base Model Params: 354.71M}}\\
						Base & 2.779 & 65.13 & 37.73 & 32.40 & 78.60 & 56.32 & 29.77& 32.76& 51.14& 47.98 \\
						MoE    & \textbf{2.585} & 68.93 & \textbf{45.33} & 33.20 & 85.40 & 63.16 & 29.77& 34.64& 51.38& 51.48 \\
						2-Loop     & 2.631 & 67.90 & 44.06 & 34.60 & 84.60 & 64.56 & 28.76& 35.46& 51.78& 51.47 \\
						4-Loop      & 2.625 & 68.44 & 44.09 & 34.60 & 84.80 & 64.56 & \textbf{30.77}& 36.53& 52.01& 51.98 \\
						6-Loop      & 2.623 & 67.16 & 44.06 & \textbf{36.00} & 84.30 & 64.04 & 30.77& 35.95& 53.20& 51.94 \\
						VersatileFFN  & 2.617 & \textbf{69.10} & 43.95 & 35.00 & \textbf{85.70} & \textbf{64.56} & 30.10& \textbf{36.69}& \textbf{53.51}& \textbf{52.33} \\
						\midrule
						\multicolumn{11}{c}{\cellcolor[HTML]{EFEFEF}\textit{Base Model Params: 720.81M}}\\
						Base & 2.519 & 69.48 & 47.69 & 37.20 & 86.40 & 65.61 & 34.11& 35.14& 55.01& 53.83 \\
						MoE    & \textbf{2.411} & \textbf{72.31} & \textbf{54.47} & 38.00 & 89.00 & 67.72 & 34.45& 37.10& 53.91& 55.87 \\
						2-Loop     & 2.448 & 71.33 & 52.99 & 37.00 & 88.20 & 68.42 & 35.45& 37.84& 55.41& 55.83 \\
						4-Loop      & 2.441 & 71.87 & 53.43 & 37.00 & 88.60 & 68.25 & \textbf{36.45}& 39.39& 55.64& 56.33 \\
						6-Loop      & 2.430 & 71.98 & 53.96 & \textbf{38.80} & \textbf{89.30} & 69.47 & 34.45& 38.98& 55.49& 56.55 \\
						VersatileFFN  & 2.430 & 72.03 & 53.44 & 38.00 & 88.90 & \textbf{71.05} & 36.12& \textbf{40.79}& \textbf{55.88}& \textbf{57.03} \\
						\midrule
						\multicolumn{11}{c}{\cellcolor[HTML]{EFEFEF}\textit{Base Model Params: 1.21B}}\\
						Base & 2.385 & 73.07 & 54.18 & 38.00 & 88.90 & 69.12 & 35.12& 40.05& 55.49& 56.74 \\
						MoE    & \textbf{2.278} & 74.21 & 59.37 & 41.00 & 91.10 & 70.35 & 41.14& 41.20& 58.80& 59.65 \\
						2-Loop     & 2.328 & 74.59 & 59.08 & 39.60 & 91.10 & 71.93 & 41.47& 40.70& 57.38& 59.48 \\
						4-Loop      & 2.322 & 73.97 & 59.41 & 40.70 & 91.50 & 72.86 & \textbf{41.51}& 42.86& 57.87& 60.09 \\
						6-Loop      & 2.320 & \textbf{74.76} & 59.86 & \textbf{41.20} & 91.40 & 71.58 & 40.47& 41.93& \textbf{59.19} & 60.05 \\
						VersatileFFN  & 2.319 & 74.65 & \textbf{60.17} & 40.80 & \textbf{91.90} & \textbf{73.16} & 41.14& \textbf{43.73}& 58.17 & \textbf{60.47} \\
						\bottomrule
					\end{tabular}
				\end{sc}
			\end{small}
		\end{center}
		\vskip -0.2in
	\end{table*}

	\begin{table}[h]
	\caption{Efficiency comparison for various methods.}
	\label{table:efficiency}
	\vspace{-1em}
	\begin{center}
		\begin{small}
			\begin{sc}
				\begin{tabular}{c|c|c|c}
					\toprule
					Method  & Avg.& Params (B) &  FLOPs (G)\\
					\midrule
					\rowcolor{gray!10}  Base & 47.98 & 0.35 & 0.38 \\
					MoE   & 51.48  & 0.54 & 0.47 \\
					2-Loop    & 51.47  & 0.35 & 0.75 \\
					4-Loop     & 51.98  & 0.35 & 1.50 \\
					6-Loop     & 51.94  & 0.35 & 2.25 \\
					VersatileFFN  & 52.33  & 0.35 & 1.24 \\
					\midrule
					\rowcolor{gray!10} Base & 53.83 & 0.72 & 0.85 \\
					MoE   & 55.87  & 1.15 & 1.06 \\
					2-Loop    & 55.83  & 0.72 & 1.70 \\
					4-Loop     & 56.33  & 0.72 & 3.40 \\
					6-Loop     & 56.55  & 0.72 & 5.10 \\
					VersatileFFN  & 57.03  & 0.72 & 2.59 \\
					\midrule
					\rowcolor{gray!10} Base & 56.74 & 1.21 & 1.51 \\
					MoE   & 59.65  & 1.97 & 1.89 \\
					2-Loop    & 59.48  & 1.21 & 3.02 \\
					4-Loop     & 60.09  & 1.21 & 6.04 \\
					6-Loop     & 60.05  & 1.21 & 9.06 \\
					VersatileFFN  & 60.47  & 1.21 & 4.69 \\
					\bottomrule
				\end{tabular}
			\end{sc}
		\end{small}
	\end{center}
	\vskip -0.2in
\end{table}
	
	\noindent\textbf{Versatile Setup of Width and Depth. } 
	To determine the optimal configuration for the VersatileFFN architecture, specifically the number of active experts and total experts in the width (\ie $k$ and $N$ in Eq.~\ref{eq:y_moe}) and the maximum loops in the depth (\ie $L_{\max}$ in Eq.~\ref{eq:y_loop}), we conduct ablation studies using the 354M model trained on 40B tokens. In these experiments, we maintained one shared expert for the width-versatile pathway, while disabling virtual experts within the depth-versatile pathway. The training loss and average accuracy across eight academic benchmarks during training is illustrated in Figure~\ref{fig:exp_settting}. For width-versatile, the losses (and average accuracies) for varying expert configurations, specifically 4-choose-1, 8-choose-2, 8-choose-4, and 16-choose-4 are 2.655 (50.16\%), 2.633 (50.98\%), 2.633 (50.96\%), and 2.634 (50.22\%), respectively. We select the 8-choose-2 setting as our primary experimental configuration due to its higher performance. For depth-versatile, when the number of loops is 1, 2, 4, and 6, the final training losses are 2.654, 2.631, 2.625, and 2.623, respectively. The average accuracies are 50.09\%, 51.47\%, 51.98\%, and 51.94\%. It can be seen that 4 loops can achieve the best balance between accuracy and computation. Therefore, we set the maximum number of loops to 4 in our main experiment.

	\subsection{Evaluation Results}
	To rigorously evaluate the efficacy of our proposed method, we conduct continued pre-training on a pre-trained base model for one additional epoch, utilizing the identical corpus. We benchmark our approach against the following methods:
	1) Mixture-of-Experts (MoE): A sparse architecture that inherits the configuration of the dense base model but incorporates additional 8 small experts with activating top-2 experts during the forward pass.
	2)  $k$-Loop: A variant of the dense architecture that retains the original architecture but introduces recurrence by iteratively applying the FFN $k$ times at each layer.
	To ensure a fair comparison, both methods also share the experimental setting of continued pre-training on the same pre-trained base model.
	
	\noindent\textbf{Efficiency Comparison. } 
	We first provide an analysis of the efficiency in Table~\ref{table:efficiency}, including the average accuracy, model parameters, and FLOPs of the FFN part. We calculate the FLOPs as 1 token input and 1 token output. Since the same architecture is maintained, the number of parameters for methods $k$-Loop is consistent with that of the base model. The proposed VersatileFFN method, however, designs a router and loop predictor, which introduces negligible parameters. The MoE method, on the other hand, introduces real small experts on top of the base model, leading to a significant increase in parameters. For FFN FLOPs, the computational cost of $k$-Loop method is clearly $k$ times that of base. We calculate the FLOPs of VersatileFFN based on the inference statistics of the ARC-c dataset. Specifically, we calculate the average loops of all layers $N_{mean}$ and the proportion $P$ of loops not equal to the maximum loop, and then compute the FLOPs by $Base \times N_{mean} + (MoE - Base) \times P$. 
	
	\noindent\textbf{Benchmark Accuracy. } 
	We report the detailed zero-shot performance of VersatileFFN against other methods in Table~\ref{table:main_result}. From the results, VersatileFFN consistently achieves the highest average accuracy across both model scales. Specifically, on the 1.21B scale, our method attains an average accuracy of 60.47\%, outperforming the strong MoE baseline (59.65\%) and the 6-Loop method (60.05\%). 
	It is worth noting that while MoE achieves the lowest loss (2.278 vs. 2.319 for VersatileFFN), this advantage does not translate into accuracy. VersatileFFN demonstrates more robust generalization capabilities, particularly on reasoning-intensive tasks such as ARC-e and COMM, where it leads by a significant margin (+2.81\% over MoE on ARC-e). 
	
	As shown in Table~\ref{table:efficiency}, a key advantage of VersatileFFN is its parameter efficiency. Unlike MoE, which significantly increase the total parameters, for example, from 1.21B to 1.97B (+63\%), VersatileFFN introduces negligible parameter overhead.
	Regarding computational complexity, VersatileFFN is significantly more efficient than the 4-Loop and 6-Loop methods. For instance, at the 350M scale, VersatileFFN requires approximately 45\% fewer FLOPs than the 6-Loop method  while achieving superior accuracy. This indicates that VersatileFFN allocates computational resources more effectively than simply stacking recurrent loops, offering a compelling balance between model size, training cost, and downstream performance.
	
	\subsection{Visualized Analysis}
\begin{figure*}[h]
		\begin{center}
			\centering
			\includegraphics[width=0.9\linewidth]{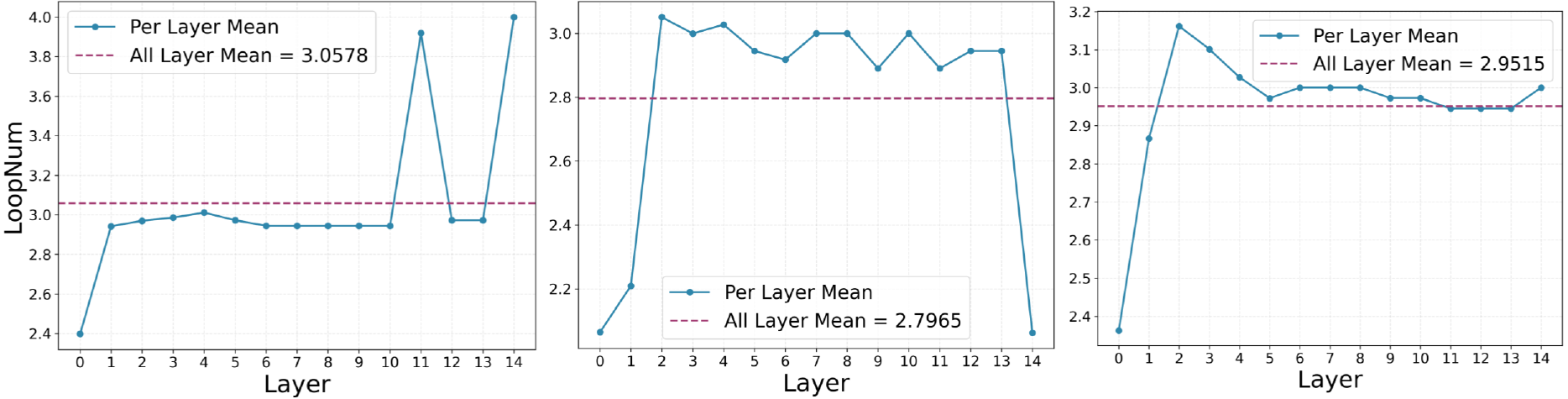}
			\caption{The average predicted loops per layer on ARC-c dataset. From left to right: 354M, 720M, 1.21B model.}
			\label{fig:loop}
		\end{center}
		\vspace{-1.0em}
	\end{figure*}

	\noindent\textbf{Actual loops. } 
	We first visualize the average predicted loop counts per layer on the ARC-c dataset in Figure~\ref{fig:loop}. The smaller 354M model exhibits a late-stage allocation strategy, dedicating the highest computational budget to the final layers (specifically layers 11–14). The medium-sized 720M model transitions to a middle-heavy distribution, concentrating recurrence primarily within the intermediate layers while tapering off at the boundaries. In contrast, the 1.21B model displays a front-loaded profile: the recurrence intensity peaks sharply at Layer 2 and subsequently stabilizes into a consistent plateau for the remainder of the network. 
	
\begin{figure}[t]
		\begin{center}
			\centering
			\includegraphics[width=0.9\linewidth]{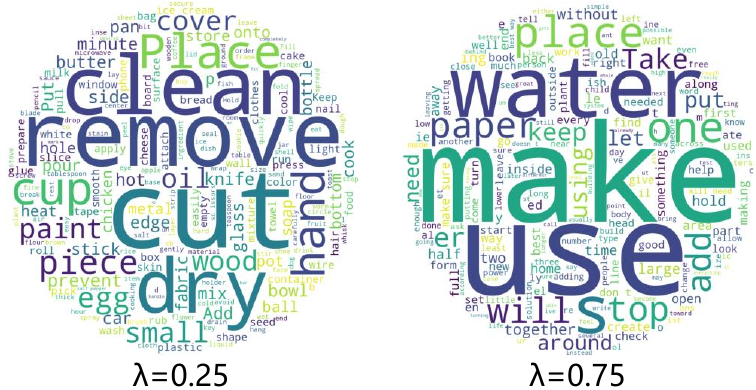}
			\caption{Word cloud for various gating coefficients on ARC-c.}
			\label{fig:wordcloud}
		\end{center}
		\vspace{-2.0em}
	\end{figure}

	\noindent\textbf{Word cloud. } 
	We then analyze the word cloud of 354M model at Layer\_0 on ARC-c in Figure~\ref{fig:wordcloud}. A lower gating coefficients $\lambda$ indicates more loops. The words in the left is dominated by specific action verbs and similar words such as \textit{clean}, \textit{remove}, \textit{cut}, \textit{cup}. Conversely, the right consists primarily of high-frequency, generic terms like \textit{make}, \textit{use}, \textit{water}, \textit{will}. These tokens typically rely on shallow surface statistics or syntactic cues, allowing the model to capture their representations with minimal recurrence.
	
	\subsection{Ablation Studies}
	We conduct a series of ablation studies using the 354M base model on 40B tokens.
		\begin{table}[h]
		\caption{Analysis on each components in VersatileFFN.}
		\label{table:ablation0}
		\vspace{-1em}
		\begin{center}
			\begin{small}
				\begin{sc}
					\setlength{\tabcolsep}{1.0mm}
					\begin{tabular}{c|ccc|cc}
						\toprule
						Method  & Width & Depth & Gating & Loss &Acc.  \\
						\midrule
						Base &  &  & & 2.779 & 47.98  \\
						\cmidrule{1-6} 
						\multirow{4}{*}{Ours} 
						& \checkmark &  & & 2.633 & 50.98  \\
						&  & \checkmark & & 2.629 & 51.35  \\
						& \checkmark & \checkmark & & 2.618 & 52.10  \\
						& \checkmark & \checkmark & \checkmark & \textbf{2.617} & \textbf{52.33}  \\
						\bottomrule
					\end{tabular}
				\end{sc}
			\end{small}
		\end{center}
		\vspace{-1.0em}
	\end{table}

	\noindent\textbf{Analysis on Each Components. } From Table~\ref{table:ablation0}, we observe that both width-versatile and depth-versatile individually outperform the baseline.
	When the fusion gating is removed ,  we adopt simple averaging to fuse the outputs of the width and depth pathways. The combination of all modules yields the lowest loss and the highest average accuracy. This result confirms that the two branches are complementary, functioning synergistically to enhance model expressivity.

	\noindent\textbf{Various MoE experts and $k$-Loop. } 
	We further investigate the impact of loop depth and expert configuration in Table~\ref{table:ablation1}. As shown in (A), while increasing the loop count to 6 minimizes  loss, the zero-shot average accuracy peaks at 4 loops, suggesting that excessive recurrence may lead to overfitting. From (B), we observe that the (8, 2) expert configuration achieves the optimal accuracy.
	\begin{table}[h]
		\caption{Analysis of the effect of: (A) Various Loops under MoE-8-2; (B) Various experts under 4-Loop.}
		\label{table:ablation1}
		\begin{center}
			\begin{small}
				\begin{sc}
					\begin{minipage}{0.48\linewidth}
						\centering
						\centerline{(a) Effect of Loops} 
						\vspace{4pt} 
						\begin{tabular}{c|c|c}
							\toprule
							Loops & Loss & Acc.\\
							\midrule
							2 & 2.625 & 51.54 \\
							4 & 2.617 & \textbf{52.33} \\
							6 & \textbf{2.615} & 51.61 \\
							\bottomrule
						\end{tabular}
					\end{minipage}
					\hfill 
					\begin{minipage}{0.48\linewidth}
						\centering
						\centerline{(b) Effect of MoE experts}
						\vspace{4pt} 
						\begin{tabular}{c|c|c}
							\toprule
							Experts & Loss & Acc.\\
							\midrule
							8-2 & 2.617 & \textbf{52.33} \\
							8-4 & 2.617 & 52.22 \\
							16-4& 2.617 & 51.72 \\
							\bottomrule
						\end{tabular}
					\end{minipage}
				\end{sc}
			\end{small}
		\end{center}
		\vskip -0.2in
	\end{table}
	
	\noindent\textbf{Performance without Continued Pre-training. } 
	Table~\ref{table:ablation2} shows the performance trained from scratch. VersatileFFN attains an accuracy of 51.14\%, yielding a 3.16\% improvement over the Base and outperforming other methods.
	\begin{table}[h]
		\caption{Performance without continued pre-training.}
		\vskip -0.1in
		\label{table:ablation2}
		\begin{center}
			\begin{small}
				\begin{sc}
					\begin{tabular}{c|c|c}
						\toprule
						Method  & Loss& Acc.\\
						\midrule
						Base & 2.779 & 47.98 \\
						MoE & \textbf{2.639} & 50.90 \\
						4-Loop      & 2.641 & 50.81 \\
						VersatileFFN  & 2.654 & \textbf{51.14} \\
						\bottomrule
					\end{tabular}
				\end{sc}
			\end{small}
		\end{center}
		\vskip -0.2in
	\end{table}
	
	\section{Conclusion}
	In this work, we present VersatileFFN, a novel architectural paradigm that decouples model performance from parameter scaling by prioritizing computational flexibility over memory expansion. By integrating a width-versatile path for efficient computing (System 1) and a depth-versatile path for iterative reasoning (System 2), our method effectively synthesizes width and depth within a strictly constrained parameter budget. Extensive experiments demonstrate that VersatileFFN consistently outperforms other methods, validating that intelligent weight reuse and adaptive computation are viable alternatives to simply scaling model size. We hope this work encourages further exploration into "compute-heavy, memory-light" architectures, paving the way for deploying sophisticated reasoning capabilities in resource-constrained environments where memory is the primary bottleneck.

	\section*{Impact Statement}
	This paper presents work whose goal is to advance the field of Machine Learning. There are many potential societal consequences of our work, none which we feel must be specifically highlighted here.
	
	\nocite{langley00}
	
	\bibliography{versatile}

@article{kaplan2020scaling,
  title={Scaling laws for neural language models},
  author={Kaplan, Jared and McCandlish, Sam and Henighan, Tom and Brown, Tom B and Chess, Benjamin and Child, Rewon and Gray, Scott and Radford, Alec and Wu, Jeffrey and Amodei, Dario},
  journal={arXiv preprint arXiv:2001.08361},
  year={2020}
}

@article{hoffmann2022training,
  title={Training compute-optimal large language models},
  author={Hoffmann, Jordan and Borgeaud, Sebastian and Mensch, Arthur and Buchatskaya, Elena and Cai, Trevor and Rutherford, Eliza and Casas, Diego de Las and Hendricks, Lisa Anne and Welbl, Johannes and Clark, Aidan and others},
  journal={arXiv preprint arXiv:2203.15556},
  year={2022}
}

@inproceedings{frantar2023sparsegpt,
  title={Sparsegpt: Massive language models can be accurately pruned in one-shot},
  author={Frantar, Elias and Alistarh, Dan},
  booktitle={International conference on machine learning},
  pages={10323--10337},
  year={2023},
  organization={PMLR}
}

@inproceedings{xiao2023smoothquant,
  title={Smoothquant: Accurate and efficient post-training quantization for large language models},
  author={Xiao, Guangxuan and Lin, Ji and Seznec, Mickael and Wu, Hao and Demouth, Julien and Han, Song},
  booktitle={International conference on machine learning},
  pages={38087--38099},
  year={2023},
  organization={PMLR}
}

@article{fu2025eaquant,
  title={EAQuant: Enhancing Post-Training Quantization for MoE Models via Expert-Aware Optimization},
  author={Fu, Zhongqian and Ding, Ning and Han, Kai and Yu, Xianzhi and Li, Xiaosong and Chen, Xinghao and Tang, Yehui and Wang, Yunhe},
  journal={arXiv preprint arXiv:2506.13329},
  year={2025}
}

@article{chen2025pangu,
  title={Pangu Light: Weight Re-Initialization for Pruning and Accelerating LLMs},
  author={Chen, Hanting and Qin, Jiarui and Guo, Jialong and Yuan, Tao and Yin, Yichun and Zhen, Huiling and Wang, Yasheng and Li, Jinpeng and Meng, Xiaojun and Zhang, Meng and others},
  journal={arXiv preprint arXiv:2505.20155},
  year={2025}
}

@inproceedings{lu2025flrc,
  title={FLRC: Fine-grained Low-Rank Compressor for Efficient LLM Inference},
  author={Lu, Yu-Chen and Chen, Chong-Yan and Chang, Chi-Chih and Hu, Yu-Fang and Wu, Kai-Chiang},
  booktitle={Proceedings of the 2025 Conference on Empirical Methods in Natural Language Processing},
  pages={14956--14966},
  year={2025}
}

@book{kahneman2011thinking,
  title={Thinking, fast and slow},
  author={Kahneman, Daniel},
  year={2011},
  publisher={macmillan}
}

@article{fedus2022switch,
  title={Switch transformers: Scaling to trillion parameter models with simple and efficient sparsity},
  author={Fedus, William and Zoph, Barret and Shazeer, Noam},
  journal={Journal of Machine Learning Research},
  volume={23},
  number={120},
  pages={1--39},
  year={2022}
}

@article{jang2016categorical,
  title={Categorical reparameterization with gumbel-softmax},
  author={Jang, Eric and Gu, Shixiang and Poole, Ben},
  journal={arXiv preprint arXiv:1611.01144},
  year={2016}
}

@article{jacobs1991adaptive,
  title={Adaptive mixtures of local experts},
  author={Jacobs, Robert A and Jordan, Michael I and Nowlan, Steven J and Hinton, Geoffrey E},
  journal={Neural computation},
  volume={3},
  number={1},
  pages={79--87},
  year={1991},
  publisher={MIT Press}
}

@article{jordan1994hierarchical,
  title={Hierarchical mixtures of experts and the EM algorithm},
  author={Jordan, Michael I and Jacobs, Robert A},
  journal={Neural computation},
  volume={6},
  number={2},
  pages={181--214},
  year={1994},
  publisher={MIT Press}
}

@article{shazeer2017outrageously,
  title={Outrageously large neural networks: The sparsely-gated mixture-of-experts layer},
  author={Shazeer, Noam and Mirhoseini, Azalia and Maziarz, Krzysztof and Davis, Andy and Le, Quoc and Hinton, Geoffrey and Dean, Jeff},
  journal={arXiv preprint arXiv:1701.06538},
  year={2017}
}

@article{lepikhin2020gshard,
  title={Gshard: Scaling giant models with conditional computation and automatic sharding},
  author={Lepikhin, Dmitry and Lee, HyoukJoong and Xu, Yuanzhong and Chen, Dehao and Firat, Orhan and Huang, Yanping and Krikun, Maxim and Shazeer, Noam and Chen, Zhifeng},
  journal={arXiv preprint arXiv:2006.16668},
  year={2020}
}

@article{jiang2024mixtral,
  title={Mixtral of experts},
  author={Jiang, Albert Q and Sablayrolles, Alexandre and Roux, Antoine and Mensch, Arthur and Savary, Blanche and Bamford, Chris and Chaplot, Devendra Singh and Casas, Diego de las and Hanna, Emma Bou and Bressand, Florian and others},
  journal={arXiv preprint arXiv:2401.04088},
  year={2024}
}

@article{dai2024deepseekmoe,
  title={Deepseekmoe: Towards ultimate expert specialization in mixture-of-experts language models},
  author={Dai, Damai and Deng, Chengqi and Zhao, Chenggang and Xu, RX and Gao, Huazuo and Chen, Deli and Li, Jiashi and Zeng, Wangding and Yu, Xingkai and Wu, Yu and others},
  journal={arXiv preprint arXiv:2401.06066},
  year={2024}
}

@article{yang2025qwen3,
  title={Qwen3 technical report},
  author={Yang, An and Li, Anfeng and Yang, Baosong and Zhang, Beichen and Hui, Binyuan and Zheng, Bo and Yu, Bowen and Gao, Chang and Huang, Chengen and Lv, Chenxu and others},
  journal={arXiv preprint arXiv:2505.09388},
  year={2025}
}

@article{wang2024remoe,
  title={Remoe: Fully differentiable mixture-of-experts with relu routing},
  author={Wang, Ziteng and Zhu, Jun and Chen, Jianfei},
  journal={arXiv preprint arXiv:2412.14711},
  year={2024}
}

@article{jin2024moe++,
  title={Moe++: Accelerating mixture-of-experts methods with zero-computation experts},
  author={Jin, Peng and Zhu, Bo and Yuan, Li and Yan, Shuicheng},
  journal={arXiv preprint arXiv:2410.07348},
  year={2024}
}

@article{huang2024harder,
  title={Harder tasks need more experts: Dynamic routing in moe models},
  author={Huang, Quzhe and An, Zhenwei and Zhuang, Nan and Tao, Mingxu and Zhang, Chen and Jin, Yang and Xu, Kun and Chen, Liwei and Huang, Songfang and Feng, Yansong},
  journal={arXiv preprint arXiv:2403.07652},
  year={2024}
}

@article{comanici2025gemini,
  title={Gemini 2.5: Pushing the frontier with advanced reasoning, multimodality, long context, and next generation agentic capabilities},
  author={Comanici, Gheorghe and Bieber, Eric and Schaekermann, Mike and Pasupat, Ice and Sachdeva, Noveen and Dhillon, Inderjit and Blistein, Marcel and Ram, Ori and Zhang, Dan and Rosen, Evan and others},
  journal={arXiv preprint arXiv:2507.06261},
  year={2025}
}

@article{team2025kimi,
  title={Kimi k2: Open agentic intelligence},
  author={Team, Kimi and Bai, Yifan and Bao, Yiping and Chen, Guanduo and Chen, Jiahao and Chen, Ningxin and Chen, Ruijue and Chen, Yanru and Chen, Yuankun and Chen, Yutian and others},
  journal={arXiv preprint arXiv:2507.20534},
  year={2025}
}

@article{zhao2025towards,
  title={Towards a Comprehensive Scaling Law of Mixture-of-Experts},
  author={Zhao, Guoliang and Fu, Yuhan and Li, Shuaipeng and Sun, Xingwu and Xie, Ruobing and Wang, An and Han, Weidong and Yang, Zhen and Sun, Weixuan and Zhang, Yudong and others},
  journal={arXiv preprint arXiv:2509.23678},
  year={2025}
}

@article{huang2023lorahub,
  title={Lorahub: Efficient cross-task generalization via dynamic lora composition},
  author={Huang, Chengsong and Liu, Qian and Lin, Bill Yuchen and Pang, Tianyu and Du, Chao and Lin, Min},
  journal={arXiv preprint arXiv:2307.13269},
  year={2023}
}

@article{feng2024mixture,
  title={Mixture-of-loras: An efficient multitask tuning for large language models},
  author={Feng, Wenfeng and Hao, Chuzhan and Zhang, Yuewei and Han, Yu and Wang, Hao},
  journal={arXiv preprint arXiv:2403.03432},
  year={2024}
}

@article{wu2024mixture,
  title={Mixture of lora experts},
  author={Wu, Xun and Huang, Shaohan and Wei, Furu},
  journal={arXiv preprint arXiv:2404.13628},
  year={2024}
}

@article{han2025moragent,
  title={MoRAgent: Parameter Efficient Agent Tuning with Mixture-of-Roles},
  author={Han, Jing and Yan, Binwei and Guo, Tianyu and Bai, Zheyuan and Zheng, Mengyu and Chen, Hanting and Nie, Ying},
  journal={arXiv preprint arXiv:2512.21708},
  year={2025}
}

@article{li2022branch,
  title={Branch-train-merge: Embarrassingly parallel training of expert language models},
  author={Li, Margaret and Gururangan, Suchin and Dettmers, Tim and Lewis, Mike and Althoff, Tim and Smith, Noah A and Zettlemoyer, Luke},
  journal={arXiv preprint arXiv:2208.03306},
  year={2022}
}

@article{zhao2024hypermoe,
  title={Hypermoe: Towards better mixture of experts via transferring among experts},
  author={Zhao, Hao and Qiu, Zihan and Wu, Huijia and Wang, Zili and He, Zhaofeng and Fu, Jie},
  journal={arXiv preprint arXiv:2402.12656},
  year={2024}
}

@article{dong2024stbllm,
  title={Stbllm: Breaking the 1-bit barrier with structured binary llms},
  author={Dong, Peijie and Li, Lujun and Zhong, Yuedong and Du, Dayou and Fan, Ruibo and Chen, Yuhan and Tang, Zhenheng and Wang, Qiang and Xue, Wei and Guo, Yike and others},
  journal={arXiv preprint arXiv:2408.01803},
  year={2024}
}

@article{huang2024mixture,
  title={Mixture Compressor for Mixture-of-Experts LLMs Gains More},
  author={Huang, Wei and Liao, Yue and Liu, Jianhui and He, Ruifei and Tan, Haoru and Zhang, Shiming and Li, Hongsheng and Liu, Si and Qi, Xiaojuan},
  journal={arXiv preprint arXiv:2410.06270},
  year={2024}
}

@article{zhou2025floe,
  title={FloE: On-the-Fly MoE Inference on Memory-constrained GPU},
  author={Zhou, Yuxin and Li, Zheng and Zhang, Jun and Wang, Jue and Wang, Yiping and Xie, Zhongle and Chen, Ke and Shou, Lidan},
  journal={arXiv preprint arXiv:2505.05950},
  year={2025}
}

@article{sarkar2024revisiting,
  title={Revisiting smoe language models by evaluating inefficiencies with task specific expert pruning},
  author={Sarkar, Soumajyoti and Lausen, Leonard and Cevher, Volkan and Zha, Sheng and Brox, Thomas and Karypis, George},
  journal={arXiv preprint arXiv:2409.01483},
  year={2024}
}

@article{lu2024not,
  title={Not all experts are equal: Efficient expert pruning and skipping for mixture-of-experts large language models},
  author={Lu, Xudong and Liu, Qi and Xu, Yuhui and Zhou, Aojun and Huang, Siyuan and Zhang, Bo and Yan, Junchi and Li, Hongsheng},
  journal={arXiv preprint arXiv:2402.14800},
  year={2024}
}

@article{dehghani2018universal,
  title={Universal transformers},
  author={Dehghani, Mostafa and Gouws, Stephan and Vinyals, Oriol and Uszkoreit, Jakob and Kaiser, {\L}ukasz},
  journal={arXiv preprint arXiv:1807.03819},
  year={2018}
}

@article{lan2019albert,
  title={Albert: A lite bert for self-supervised learning of language representations},
  author={Lan, Zhenzhong and Chen, Mingda and Goodman, Sebastian and Gimpel, Kevin and Sharma, Piyush and Soricut, Radu},
  journal={arXiv preprint arXiv:1909.11942},
  year={2019}
}

@article{gao2024expressive,
  title={On the expressive power of a variant of the looped transformer},
  author={Gao, Yihang and Zheng, Chuanyang and Xie, Enze and Shi, Han and Hu, Tianyang and Li, Yu and Ng, Michael K and Li, Zhenguo and Liu, Zhaoqiang},
  journal={CoRR},
  year={2024}
}

@article{bae2025mixture,
  title={Mixture-of-recursions: Learning dynamic recursive depths for adaptive token-level computation},
  author={Bae, Sangmin and Kim, Yujin and Bayat, Reza and Kim, Sungnyun and Ha, Jiyoun and Schuster, Tal and Fisch, Adam and Harutyunyan, Hrayr and Ji, Ziwei and Courville, Aaron and others},
  journal={arXiv preprint arXiv:2507.10524},
  year={2025}
}

@inproceedings{giannou2023looped,
  title={Looped transformers as programmable computers},
  author={Giannou, Angeliki and Rajput, Shashank and Sohn, Jy-yong and Lee, Kangwook and Lee, Jason D and Papailiopoulos, Dimitris},
  booktitle={International Conference on Machine Learning},
  pages={11398--11442},
  year={2023},
  organization={PMLR}
}

@article{fan2024looped,
  title={Looped transformers for length generalization},
  author={Fan, Ying and Du, Yilun and Ramchandran, Kannan and Lee, Kangwook},
  journal={arXiv preprint arXiv:2409.15647},
  year={2024}
}

@article{saunshi2025reasoning,
  title={Reasoning with latent thoughts: On the power of looped transformers},
  author={Saunshi, Nikunj and Dikkala, Nishanth and Li, Zhiyuan and Kumar, Sanjiv and Reddi, Sashank J},
  journal={arXiv preprint arXiv:2502.17416},
  year={2025}
}

@article{wang2025hierarchical,
  title={Hierarchical Reasoning Model},
  author={Wang, Guan and Li, Jin and Sun, Yuhao and Chen, Xing and Liu, Changling and Wu, Yue and Lu, Meng and Song, Sen and Yadkori, Yasin Abbasi},
  journal={arXiv preprint arXiv:2506.21734},
  year={2025}
}

@article{jolicoeur2025less,
  title={Less is more: Recursive reasoning with tiny networks},
  author={Jolicoeur-Martineau, Alexia},
  journal={arXiv preprint arXiv:2510.04871},
  year={2025}
}

@article{zhu2025scaling,
  title={Scaling Latent Reasoning via Looped Language Models},
  author={Zhu, Rui-Jie and Wang, Zixuan and Hua, Kai and Zhang, Tianyu and Li, Ziniu and Que, Haoran and Wei, Boyi and Wen, Zixin and Yin, Fan and Xing, He and others},
  journal={arXiv preprint arXiv:2510.25741},
  year={2025}
}

@article{raposo2024mixture,
  title={Mixture-of-depths: Dynamically allocating compute in transformer-based language models},
  author={Raposo, David and Ritter, Sam and Richards, Blake and Lillicrap, Timothy and Humphreys, Peter Conway and Santoro, Adam},
  journal={arXiv preprint arXiv:2404.02258},
  year={2024}
}

@article{gong2025makes,
  title={What Makes Looped Transformers Perform Better Than Non-Recursive Ones (Provably)},
  author={Gong, Zixuan and Teng, Jiaye and Liu, Yong},
  journal={arXiv preprint arXiv:2510.10089},
  year={2025}
}

@article{gatmiry2024can,
  title={Can looped transformers learn to implement multi-step gradient descent for in-context learning?},
  author={Gatmiry, Khashayar and Saunshi, Nikunj and Reddi, Sashank J and Jegelka, Stefanie and Kumar, Sanjiv},
  journal={arXiv preprint arXiv:2410.08292},
  year={2024}
}

@article{merrill2025little,
  title={A little depth goes a long way: The expressive power of log-depth transformers},
  author={Merrill, William and Sabharwal, Ashish},
  journal={arXiv preprint arXiv:2503.03961},
  year={2025}
}

@article{penedo2024fineweb,
  title={The fineweb datasets: Decanting the web for the finest text data at scale},
  author={Penedo, Guilherme and Kydl{\'\i}{\v{c}}ek, Hynek and Lozhkov, Anton and Mitchell, Margaret and Raffel, Colin A and Von Werra, Leandro and Wolf, Thomas and others},
  journal={Advances in Neural Information Processing Systems},
  volume={37},
  pages={30811--30849},
  year={2024}
}

@misc{huggingfacefw_2024,
  author       = { HuggingFaceFW },
  title        = { fineweb-edu (Revision 22b0aca) },
  year         = 2024,
  url          = { https://huggingface.co/datasets/HuggingFaceFW/fineweb-edu },
  doi          = { 10.57967/hf/2497 },
  publisher    = { Hugging Face }
}

@article{olmo20242,
  title={2 OLMo 2 Furious},
  author={OLMo, Team and Walsh, Pete and Soldaini, Luca and Groeneveld, Dirk and Lo, Kyle and Arora, Shane and Bhagia, Akshita and Gu, Yuling and Huang, Shengyi and Jordan, Matt and others},
  journal={arXiv preprint arXiv:2501.00656},
  year={2024}
}

@article{zhang2019root,
  title={Root mean square layer normalization},
  author={Zhang, Biao and Sennrich, Rico},
  journal={Advances in neural information processing systems},
  volume={32},
  year={2019}
}

@inproceedings{zellers-etal-2019-hellaswag,
    title = "{H}ella{S}wag: Can a Machine Really Finish Your Sentence?",
    author = "Zellers, Rowan  and
      Holtzman, Ari  and
      Bisk, Yonatan  and
      Farhadi, Ali  and
      Choi, Yejin",
    editor = "Korhonen, Anna  and
      Traum, David  and
      M\`arquez, Llu\'\i s",
    booktitle = "Proceedings of the 57th Annual Meeting of the Association for Computational Linguistics",
    month = jul,
    year = "2019",
    address = "Florence, Italy",
    publisher = "Association for Computational Linguistics",
    url = "https://aclanthology.org/P19-1472/",
    doi = "10.18653/v1/P19-1472",
    pages = "4791--4800",
}

@article{clark2018think,
  title={Think you have solved question answering? try arc, the ai2 reasoning challenge},
  author={Clark, Peter and Cowhey, Isaac and Etzioni, Oren and Khot, Tushar and Sabharwal, Ashish and Schoenick, Carissa and Tafjord, Oyvind},
  journal={arXiv preprint arXiv:1803.05457},
  year={2018}
}

@inproceedings{clark-etal-2019-boolq,
    title = "{B}ool{Q}: Exploring the Surprising Difficulty of Natural Yes/No Questions",
    author = "Clark, Christopher  and
      Lee, Kenton  and
      Chang, Ming-Wei  and
      Kwiatkowski, Tom  and
      Collins, Michael  and
      Toutanova, Kristina",
    editor = "Burstein, Jill  and
      Doran, Christy  and
      Solorio, Thamar",
    booktitle = "Proceedings of the 2019 Conference of the North {A}merican Chapter of the Association for Computational Linguistics: Human Language Technologies, Volume 1 (Long and Short Papers)",
    month = jun,
    year = "2019",
    address = "Minneapolis, Minnesota",
    publisher = "Association for Computational Linguistics",
    url = "https://aclanthology.org/N19-1300/",
    doi = "10.18653/v1/N19-1300",
    pages = "2924--2936",
}

@inproceedings{OpenBookQA2018,
    title={Can a Suit of Armor Conduct Electricity? A New Dataset for Open Book Question Answering},
    author={Todor Mihaylov and Peter Clark and Tushar Khot and Ashish Sabharwal},
    booktitle={EMNLP},
    year={2018}
}

@inproceedings{Bisk2020,
    author = {Yonatan Bisk and Rowan Zellers and
            Ronan Le Bras and Jianfeng Gao
            and Yejin Choi},
    title = {PIQA: Reasoning about Physical Commonsense in
           Natural Language},
    booktitle = {Thirty-Fourth AAAI Conference on
               Artificial Intelligence},
    year = {2020},
}

@inproceedings{Welbl2017CrowdsourcingMC,
    title={Crowdsourcing Multiple Choice Science Questions},
    author={Johannes Welbl and Nelson F. Liu and Matt Gardner},
    booktitle={NUT@EMNLP},
    year={2017}
}

@article{sakaguchi2019winogrande,
    title={WinoGrande: An Adversarial Winograd Schema Challenge at Scale},
    author={Sakaguchi, Keisuke and Bras, Ronan Le and Bhagavatula, Chandra and Choi, Yejin},
    journal={arXiv preprint arXiv:1907.10641},
    year={2019}
}

@inproceedings{zuo2025man,
  title={L-Man: A Large Multi-modal Model Unifying Human-centric Tasks},
  author={Zuo, Jialong and Nie, Ying and Guo, Tianyu and Zhang, Huaxin and Hong, Jiahao and Sang, Nong and Gao, Changxin and Han, Kai},
  booktitle={Proceedings of the AAAI Conference on Artificial Intelligence},
  volume={39},
  number={10},
  pages={11095--11103},
  year={2025}
}

@article{nie2022redistribution,
  title={Redistribution of weights and activations for AdderNet quantization},
  author={Nie, Ying and Han, Kai and Diao, Haikang and Liu, Chuanjian and Wu, Enhua and Wang, Yunhe},
  journal={Advances in Neural Information Processing Systems},
  volume={35},
  pages={22739--22751},
  year={2022}
}

@article{touvron2023llama,
  title={Llama: Open and efficient foundation language models},
  author={Touvron, Hugo and Lavril, Thibaut and Izacard, Gautier and Martinet, Xavier and Lachaux, Marie-Anne and Lacroix, Timoth{\'e}e and Rozi{\`e}re, Baptiste and Goyal, Naman and Hambro, Eric and Azhar, Faisal and others},
  journal={arXiv preprint arXiv:2302.13971},
  year={2023}
}

@article{zhu2021dynamic,
  title={Dynamic resolution network},
  author={Zhu, Mingjian and Han, Kai and Wu, Enhua and Zhang, Qiulin and Nie, Ying and Lan, Zhenzhong and Wang, Yunhe},
  journal={Advances in Neural Information Processing Systems},
  volume={34},
  pages={27319--27330},
  year={2021}
}
	\bibliographystyle{icml2026}
	
	\newpage
	\appendix
	\onecolumn
	\section{Inference Optimization}
	To minimize latency during inference, we introduce two inference-time optimizations: Discrete Early-Exit and Conditional Parallelism. The complete inference procedure is illustrated in Algorithm~\ref{alg:versatile_infer}.
	
	\begin{algorithm}[h]
		\caption{Inference Computation of VersatileFFN}
		\label{alg:versatile_infer}
		\begin{algorithmic}[1]
			\REQUIRE Input tensor $\mathbf{X}$, Shared Weights $\mathbf{W}_{proj}, \mathbf{W}_{out}$, Max loops $L_{max}$
		    \ENSURE Output tensor $\mathbf{Y}$
		    \STATE \textbf{\textit{Controller \& Gating:}}
			\STATE \hspace{1em} Compute post-attention representation $\mathbf{H}$ via Eq.~\ref{eq:att}
			\STATE \hspace{1em} Compute loop probabilities $\mathbf{p} = \text{Softmax}(\mathbf{W}_{loop}\mathbf{H})$
			\STATE \hspace{1em} Determine discrete loop count $\hat{\ell} = \arg\max(\mathbf{p})$
			\STATE \hspace{1em} Compute difficulty-aware fusion scalar $\lambda$ via Eq.~\ref{eq:fusion_weights}
			
			\STATE \textbf{\textit{Width-Versatile Pathway (Conditional):}}
			\STATE \hspace{1em} \textbf{if} $\lambda > 0$ \textbf{then}
			\STATE \hspace{2em} Construct virtual experts by slicing shared weights Eq.~\ref{eq:index} and Eq.~\ref{eq:build_moe}
			\STATE \hspace{2em} Compute $\mathbf{Y}_{width}$ with Top-$k$ routing via Eq.~\ref{eq:y_moe}
			\STATE \hspace{1em} \textbf{else}
			\STATE \hspace{2em} $\mathbf{Y}_{width} \leftarrow 0$ \COMMENT{Prune branch for efficiency}
			\STATE \hspace{1em} \textbf{end if}
			
			\STATE \textbf{\textit{Depth-Versatile Pathway (Discrete Early-Exit):}}
			\STATE \hspace{1em} Initialize $\mathbf{H}^{(0)} \leftarrow \mathbf{H}$
			\STATE \hspace{1em} \textbf{for} $\ell = 1$ to $\hat{\ell}$
			\STATE \hspace{2em} Update recursive hidden state $\mathbf{H}^{(\ell)}$ using full shared weights via Eq.~\ref{eq:loop}
			\STATE \hspace{1em} \textbf{end for}
			\STATE \hspace{1em} $\mathbf{Y}_{depth} \leftarrow \mathbf{H}^{(\hat{\ell})}$ \COMMENT{Hard selection of final state}
			
			\STATE \textbf{\textit{Difficulty-Aware Fusion:}}
			\STATE \hspace{1em} Compute output $\mathbf{Y}$ by fusing pathways via Eq.~\ref{eq:fusion}
			
			\STATE \textbf{Return} $\mathbf{Y}$
		\end{algorithmic}
	\end{algorithm}

	\section{Implementation Details}
	\label{appendix_0}
	\noindent\textbf{Experiments Setup. } 
	We primarily conduct experiments under a continued pre-training setting. Specifically, we initialize the model using the pre-trained checkpoint and continue training for one epoch on the original corpus. For all models, we set the training sequence length to 4,096 tokens. AdamW optimizer is employed with a weight decay of 0.1 and a gradient clipping threshold of 1.0. The learning rate follows a cosine decay schedule, beginning with a warm-up phase for the first 5\% of total steps and subsequently decaying to 10\% of the peak value. Regarding model-specific hyperparameters, we use a global batch size of 1.5M tokens with a peak learning rate of $5 \times 10^{-4}$ for the 354M model, a global batch size of 1M tokens with a peak learning rate of $4 \times 10^{-4}$ for the 720M model and a global batch size of 0.5M tokens with a peak learning rate of $3 \times 10^{-4}$ for the 1.21B model. Furthermore, for our method and traditional MoE, we incorporate an auxiliary load balancing loss with a weighting coefficient of $1 \times 10^{-5}$.
	
	\noindent\textbf{Evaluation Benchmark Setup. } 
	We evaluate our method on a comprehensive set of benchmarks: PIQA~\cite{Bisk2020}, HellaSwag~\cite{zellers-etal-2019-hellaswag}, OpenBookQA (OBQA)~\cite{OpenBookQA2018}, SciQ~\cite{Welbl2017CrowdsourcingMC}, ARC-easy (ARC-e) and ARC-challenge (ARC-c)~\cite{clark2018think}, CommonsenseQA (COMM) ~\cite{clark-etal-2019-boolq},  and Winogrande (WINO)~\cite{sakaguchi2019winogrande}. We utilize OLMES~\cite{olmo20242} to assess the performance under zero-shot setting.
	
	\section{Features of two branches } 
\begin{figure}[t]
		\begin{center}
			\centering
			\includegraphics[width=0.9\linewidth]{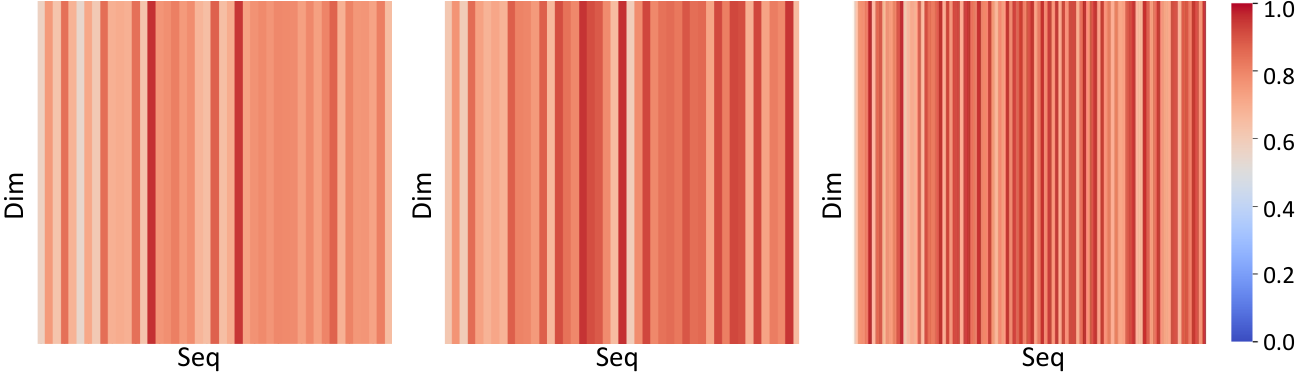}
			\caption{Heatmap of cosine similarity at Layer 0. From left to right: 354M, 720M, 1.21B model.}
			\label{fig:feature}
		\end{center}
	\end{figure}

	We analyze the features of the width-versatile and depth-versatile at Layer\_0 using a random sample from the ARC-c, as visualized in Figure~\ref{fig:feature}. Although both branches inherit the same base parameters, the output features of the two branches are not exact replicas. This demonstrates that they learn complementary representations within a globally aligned semantic space.
	
	\section{Evolution of Downstream Accuracy During Training } 
\begin{figure}[h]
		\begin{center}
			\centering
			\includegraphics[width=0.95\linewidth]{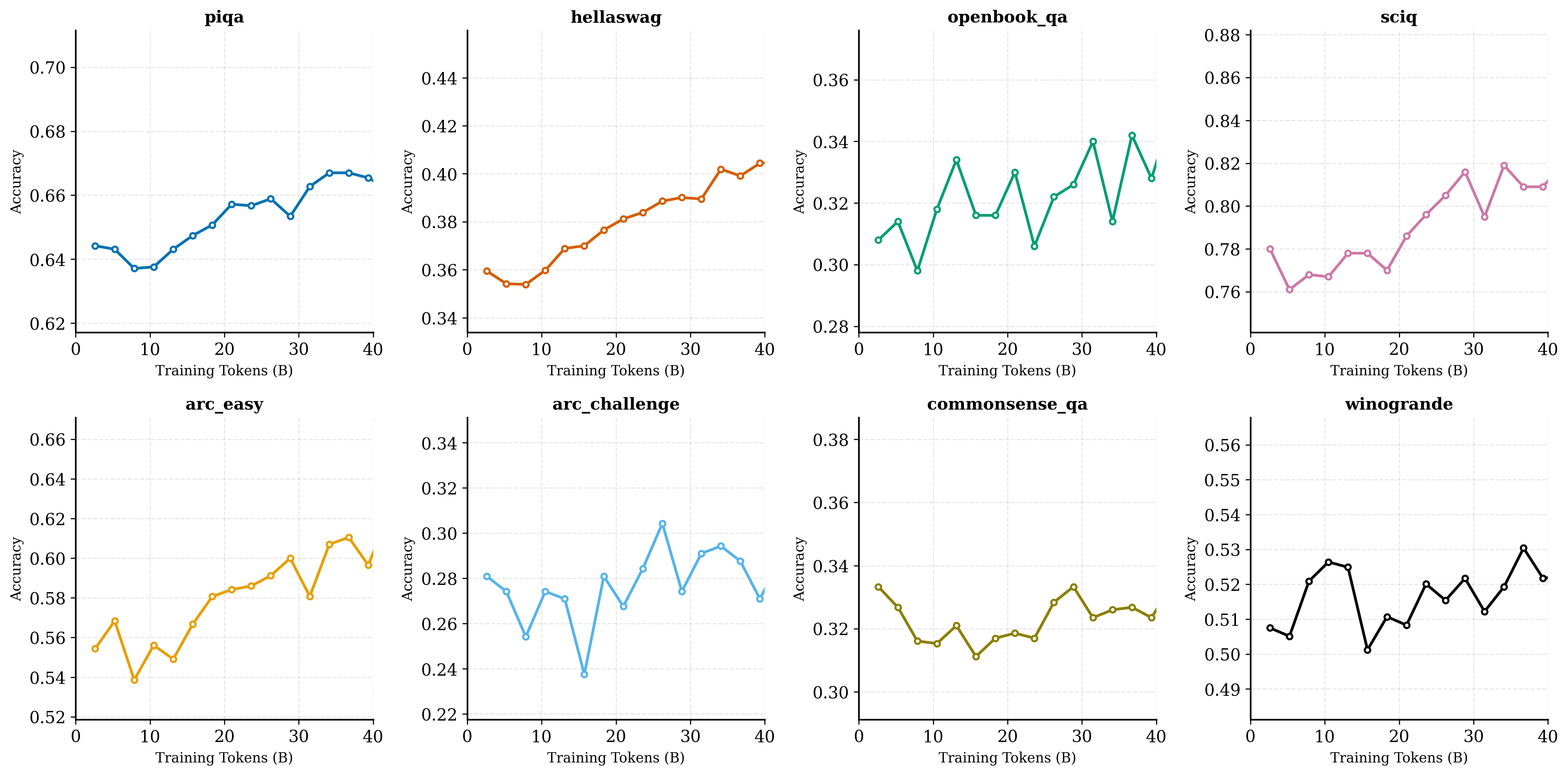}
			\caption{Accuracy of the 354M model on eight benchmarks as training progresses.}
			\label{fig:354m_acc}
		\end{center}
		\vspace{-2.0em}
	\end{figure}

\begin{figure}[h]
		\begin{center}
			\centering
			\includegraphics[width=0.95\linewidth]{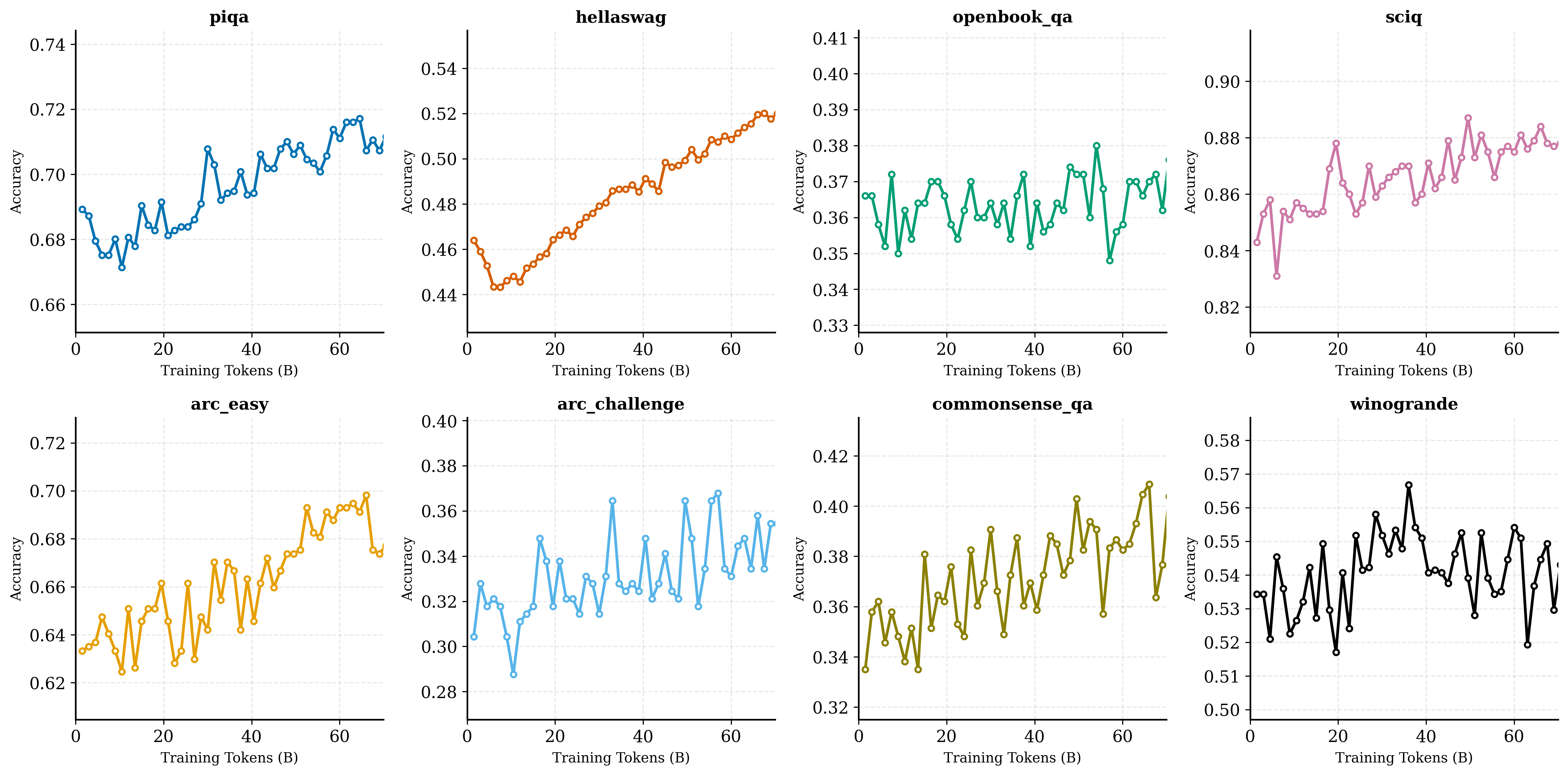}
			\caption{Accuracy of the 720M model on eight benchmarks as training progresses.}
			\label{fig:720m_acc}
		\end{center}
		\vspace{-2.0em}
	\end{figure}

\begin{figure}[t]
		\begin{center}
			\centering
			\includegraphics[width=0.95\linewidth]{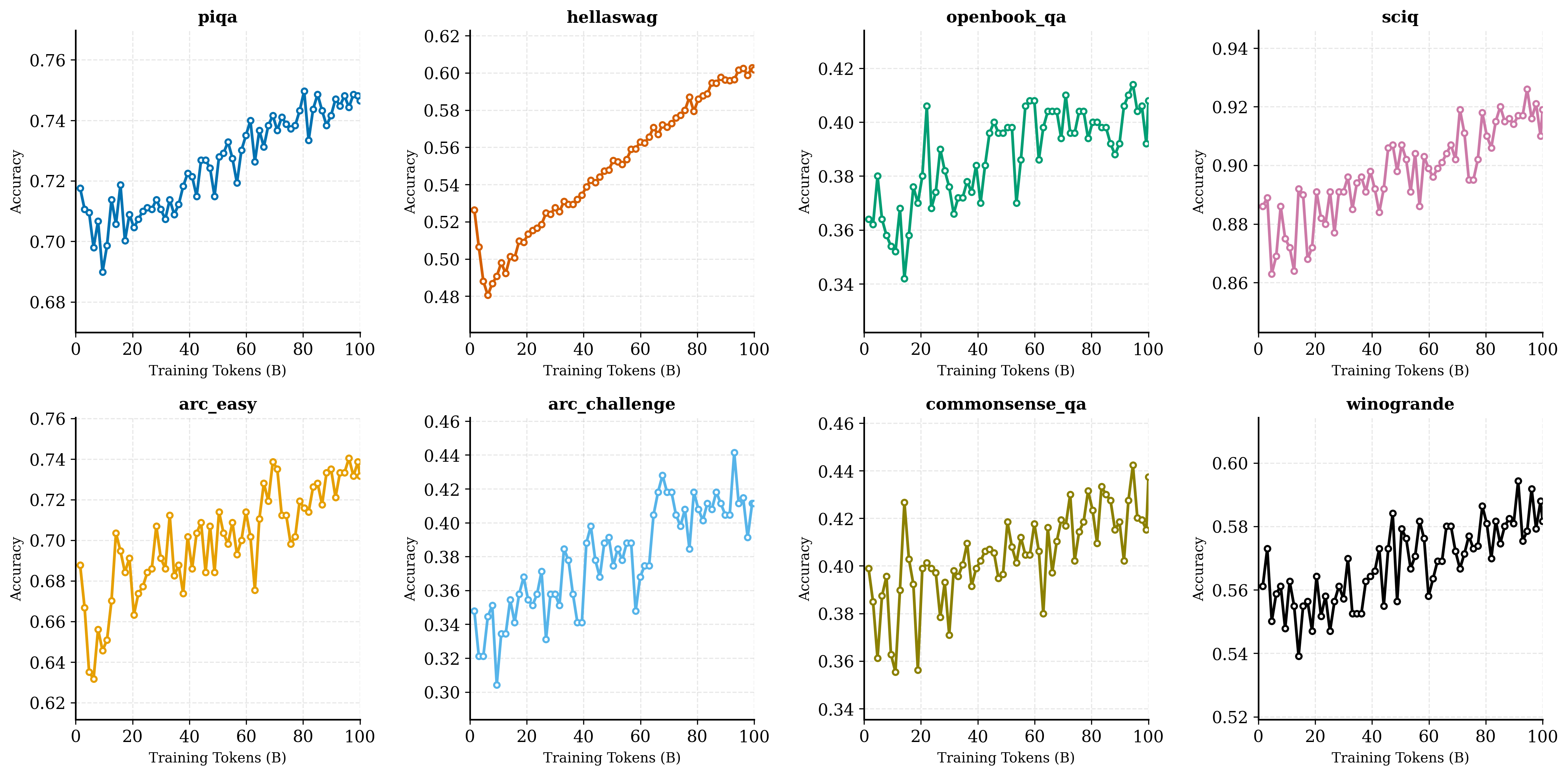}
			\caption{Accuracy of the 1.21B model on eight benchmarks as training progresses.}
			\label{fig:1b_acc}
		\end{center}
		\vspace{-2.0em}
	\end{figure}

	We illustrate the zero-shot accuracy trajectories across eight benchmarks for three distinct model scales equipped with VersatileFFN in Figure~\ref{fig:354m_acc}, Figure~\ref{fig:720m_acc} and Figure~\ref{fig:1b_acc}. We observe a consistent positive correlation between compute budget and downstream performance. As the number of training tokens increases, accuracy generally trends upward across all tasks and model sizes. Comparing the three figures reveals clear scaling benefits: the 1.21B parameter model achieves higher absolute accuracy peaks and steeper learning curves compared to its smaller counterparts. Notably, while the global trend is monotonic, the learning curves exhibit high-frequency oscillations, particularly evident in the 720M and 1.21B models on tasks such as ARC-Easy and OpenBookQA.
	
\end{document}